\documentclass[conference,a4paper]{IEEEtran}
\IEEEoverridecommandlockouts

\usepackage[hidelinks]{hyperref}
\usepackage[cmex10]{amsmath}
\usepackage{amssymb,amsfonts}
\usepackage{dblfloatfix}

\usepackage[ruled,vlined]{algorithm2e}
\usepackage{graphicx}
\graphicspath{{Figures/PDF/}{Figures/PNG/}}

\usepackage{booktabs}
\usepackage{siunitx}
\usepackage[numbers,compress]{natbib}
\usepackage{texnames}
\usepackage{bm,bbm}
\usepackage{orcidlink}

\begin{document}


\title{RemoteVAR: Autoregressive Visual Modeling for Remote Sensing Change Detection}


    \author{	\IEEEauthorblockN{Yilmaz Korkmaz}
	\IEEEauthorblockA{\textit{Johns Hopkins University}\\
		Baltimore, Maryland, USA\\
		ykorkma1@jhu.edu}
	\and
	\IEEEauthorblockN{Vishal M. Patel}
	\IEEEauthorblockA{\textit{Johns Hopkins University}\\
		Baltimore, Maryland, USA\\
		vpatel36@jhu.edu}
}

\maketitle
\begin{abstract}
Remote sensing change detection aims to localize and characterize scene changes between two time points and is central to applications such as environmental monitoring and disaster assessment. Meanwhile, visual autoregressive models (VARs) have recently shown impressive image generation capability, but their adoption for pixel-level discriminative tasks remains limited due to weak controllability, suboptimal dense prediction performance and exposure bias. We introduce RemoteVAR, a new VAR-based change detection framework that addresses these limitations by conditioning autoregressive prediction on multi-resolution fused bi-temporal features via cross-attention, and by employing an autoregressive training strategy designed specifically for change map prediction. Extensive experiments on standard change detection benchmarks show that RemoteVAR delivers consistent and significant improvements over strong diffusion-based and transformer-based baselines, establishing a competitive autoregressive alternative for remote sensing change detection. Code will be available \href{https://github.com/yilmazkorkmaz1/RemoteVAR}{\underline{here}}.
\end{abstract}


\section{Introduction}

In remote sensing, Change Detection (CD) aims to identify alterations on the Earth's surface over time by comparing satellite observations of the same area acquired at different time points \cite{cd1}. CD serves as a core component in a broad set of real-world applications, including monitoring and assessing natural disasters and climate variability \cite{climateshift_cd1,disaster_cd1,disaster_cd2}, supporting policy and urban planning decisions \cite{policy_cd1}, mapping land use and agricultural cover \cite{land_cd1,land_cd2}, and military-oriented analysis \cite{military_cd1}. In practice, the main difficulty in CD is not the definition of the task but the nuisance differences that arise between acquisitions. Multi-temporal satellite images frequently vary due to changing illumination conditions \cite{illumination_challenge1,illumination_challenge2}, misalignment from imperfect registration \cite{registration_challenge1, registration_challenge2}, sensor-specific spatial resolution \cite{resolution_challenge1}. and measurement noise \cite{noise_challenge1, noise_challenge2}. These effects can mimic real changes and lead to false detections, making robustness to cross-time inconsistencies a central challenge for reliable CD.

Motivated by the success of deep learning, many modern CD pipelines rely on deep neural networks and have reported substantial improvements across multiple benchmarks \cite{snunet,dtscn,bit,changeformer, ddpmcd, ifnet}. Representative directions include convolutional architectures \cite{snunet,fcsiam,dtscn}, transformer-based models \cite{changeformer,bit}, and more recently diffusion-based approaches \cite{ddpmcd, wen2024gcd}. 

Most supervised change detection (CD) methods are discriminative: given a bi-temporal pair (pre-change and post-change images), they directly regress a change segmentation mask in a single shot. However, CD inherently couples global reasoning with precise localization: the model must first reconcile the overall scene context across time (e.g., viewpoint, illumination, and background content) and then resolve fine-grained change boundaries. This structure makes CD naturally compatible with a coarse-to-fine prediction process that resembles how humans inspect changes, forming an initial global hypothesis and progressively refining local details. Accordingly, we formulate CD as conditional autoregressive prediction in a discrete token space, where the change map is generated stage by stage from low to high resolution.

This view is timely given the rapid progress of visual autoregressive models (VARs) \cite{tian2024visual}. Compared with diffusion-based generative approaches, VAR-style generation provides a practical efficiency advantage at inference: diffusion models typically require many iterative denoising steps, whereas VAR produces outputs in a small number of coarse-to-fine autoregressive stages. VARs have demonstrated strong generative capabilities, yet their adaptability to discriminative, pixel-level tasks has been rarely explored and has never been studied for remote sensing change detection. To bridge this gap, we propose \textbf{RemoteVAR}, a VAR-based CD framework that redesigns conditioning and training for controllable, high-accuracy change map generation. RemoteVAR produces the change map directly via coarse-to-fine autoregressive decoding conditioned on multi-resolution bi-temporal cues.

\begin{figure*}
    \centering
    \includegraphics[width=1\linewidth]{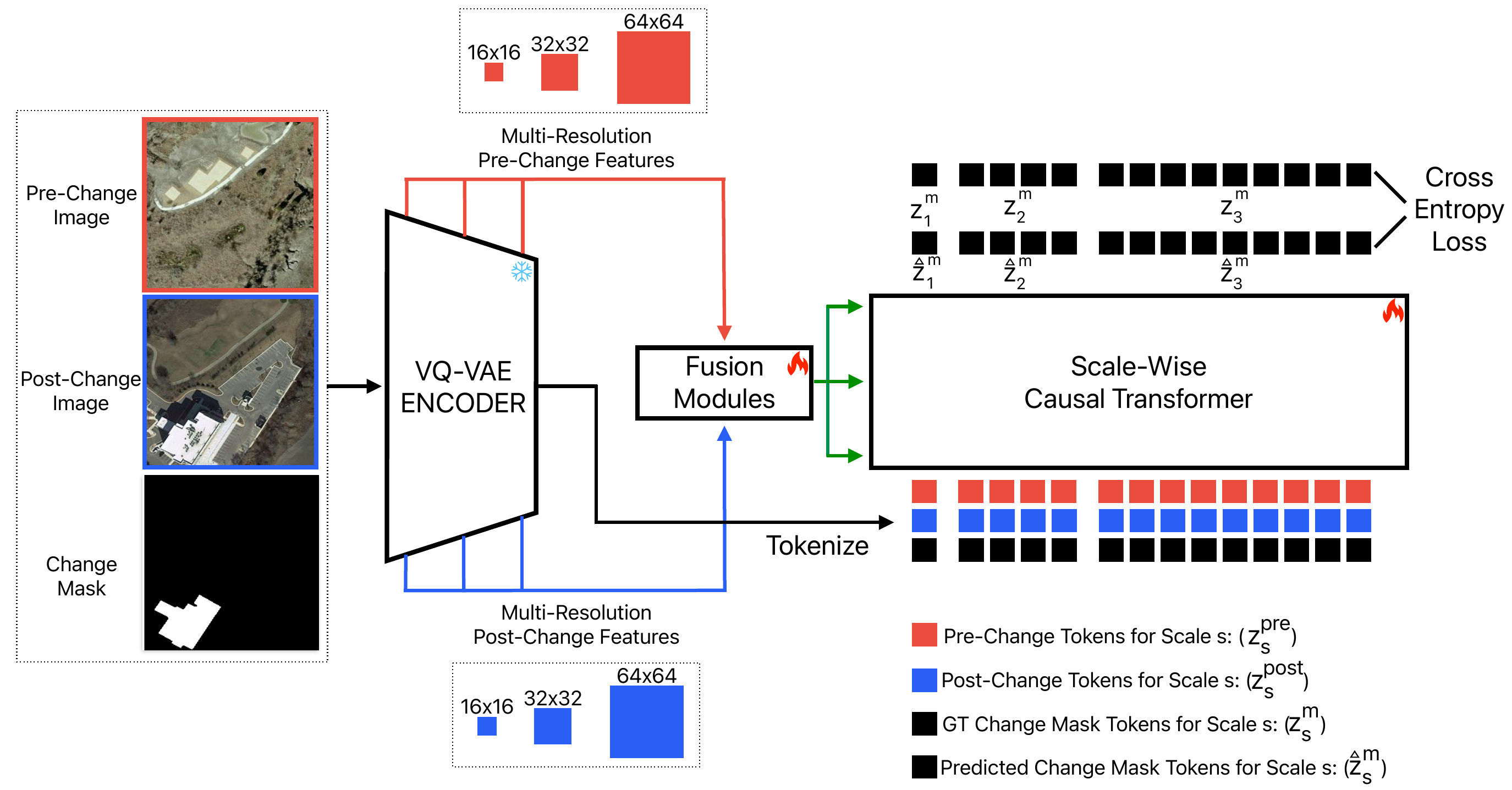}
    \caption{Overview of the RemoteVAR architecture and training pipeline. For clarity, we visualize only the first three token scales with grid sizes $1{\times}1$, $2{\times}2$, and $3{\times}3$. Pre-image, post-image, and fused feature streams are color-coded in red, blue, and green, respectively. Trainable modules are marked with a fire icon, while frozen components are marked with an ice icon.}
    \label{fig:main}
\end{figure*}

\section{Background}
Visual Autoregressive Models (VAR) \cite{tian2024visual} predict discrete visual tokens using scale-wise autoregression, generating tokens from coarse to fine resolutions to form global structure before refining details. These tokens are obtained with a residual multi-scale VQ-VAE, which constructs a tokenization scheme with varying token resolutions to encode an image into a pyramid of multi-scale tokens from coarse to fine, and quantizes latent grids into codebook indices where finer scales capture remaining details. ControlVAR \cite{li2024controlvar} extends VAR to conditional prediction by guiding token generation with additional context tokens (e.g., segmentation masks), improving controllability of the output.

\section{Methodology}

Given a bi-temporal pair consisting of a pre-change and post-change image, we perform change detection by generating discrete mask tokens rather than directly regressing pixels. We adopt the original VAR residual VQ-VAE as a frozen, scale-wise residual tokenizer with 10 token resolutions (from $\mathbf{1\times1}$ to $\mathbf{16\times16}$ across 10 scales: $s\in\{1,2,3,4,5,6,8,10,13,16\}$ with token grids $s\times s$) to map each pre/post image (and the ground-truth mask during training) into token IDs from a fixed vocabulary of size $V{=}4096$. For each scale $s$, we obtain token grids $\mathbf{z}^{pre}_s$, $\mathbf{z}^{post}_s$, and $\mathbf{z}^{m}_s$; token IDs are converted to embeddings via the shared codebook and projected to the VAR embedding dimension. To mitigate severe foreground--background imbalance, we convert binary mask tokens into a location aware RGB-coded representations following \cite{li2024controlvar} to increase token diversity, while keeping an efficient inverse mapping back to a binary mask.

We then construct a scale-wise sequence by interleaving tokens per scale as
$[\mathbf{z}^{pre}_s,\,\mathbf{z}^{post}_s,\,\mathbf{z}^{m}_s]$,
and enrich it with absolute 2D positional embeddings as well as scale embeddings so the model can distinguish pre/post/mask streams while still sharing parameters across scales. During training, we use teacher forcing and compute the loss only on mask tokens, treating pre- and post-image tokens as context for self-attention.

In contrast to ControlVAR~\cite{li2024controlvar}, which relies only on self-attention among discrete tokens for conditioning, we introduce an explicit cross-attention mechanism that injects fused continuous features into the causal transformer to improve spatial grounding for mask prediction. Specifically, beyond discrete token conditioning, we derive continuous, pixel-level feature maps directly from the VQ-VAE encoder that is already used for tokenization, removing the need for an additional feature-extraction backbone. The VQ-VAE encoder is reused as a shared-weight pre/post encoder, analogous to the Siamese encoders that are widely adopted in change detection for modeling bi-temporal inputs. To explicitly combine information across time, we apply lightweight fusion modules adapted from CMX~\cite{zhang2023cmx} to fuse the continuous features from the pre- and post-image streams into a unified conditioning representation, which is then used as the cross-attention memory, i.e., its projected embeddings serve as the keys and values, while the causal transformer token states act as queries. Unlike discrete tokens, these continuous features do not suffer from tokenization/discretization artifacts, providing crucial fine-grained information for accurate boundary localization and small-object changes. Moreover, to reduce exposure bias from teacher forcing, we randomly replace a subset of early-scale mask tokens with random codebook tokens during training, so the model learns to recover the ground-truth tokens under imperfect coarse predictions that better match autoregressive inference. The overall training procedure and architecture is illustrated in Fig.~\ref{fig:main}.

\begin{figure*}
\centering
    \includegraphics[width=.9\linewidth]{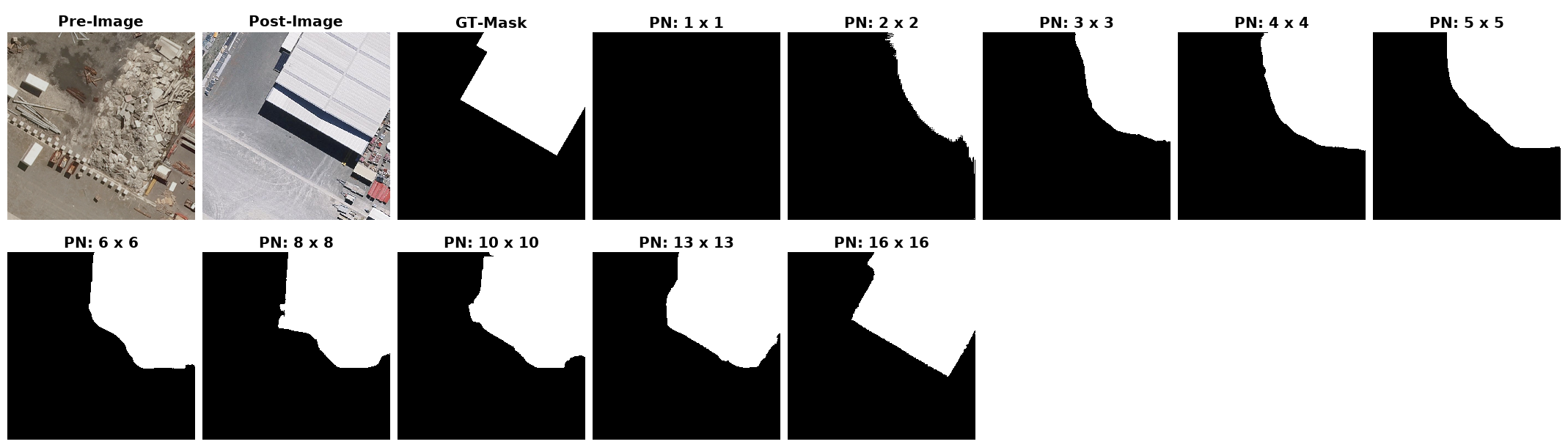}
    \caption{Scale-wise autoregressive mask generation shown across progressively finer token resolutions from $\mathbf{1\times1}$ to $\mathbf{16\times16}$.}
    \label{fig:intermediate}
\end{figure*}

At inference time, we provide $\mathbf{z}^{pre}_s$ and $\mathbf{z}^{post}_s$ via teacher forcing and autoregressively predict $\mathbf{z}^{m}_s$ in a coarse-to-fine, scale-wise manner. To transition between successive scales, we upsample the predicted mask tokens from the previous resolution to the next target grid before generation (e.g., the $10{\times}10$ prediction is upsampled to $13{\times}13$ and used as the starting context for the 9th scale), enabling progressive refinement of the change map as spatial resolution increases (see Fig. \ref{fig:intermediate}).

\begin{figure}
    \centering
    \includegraphics[width=1\linewidth]{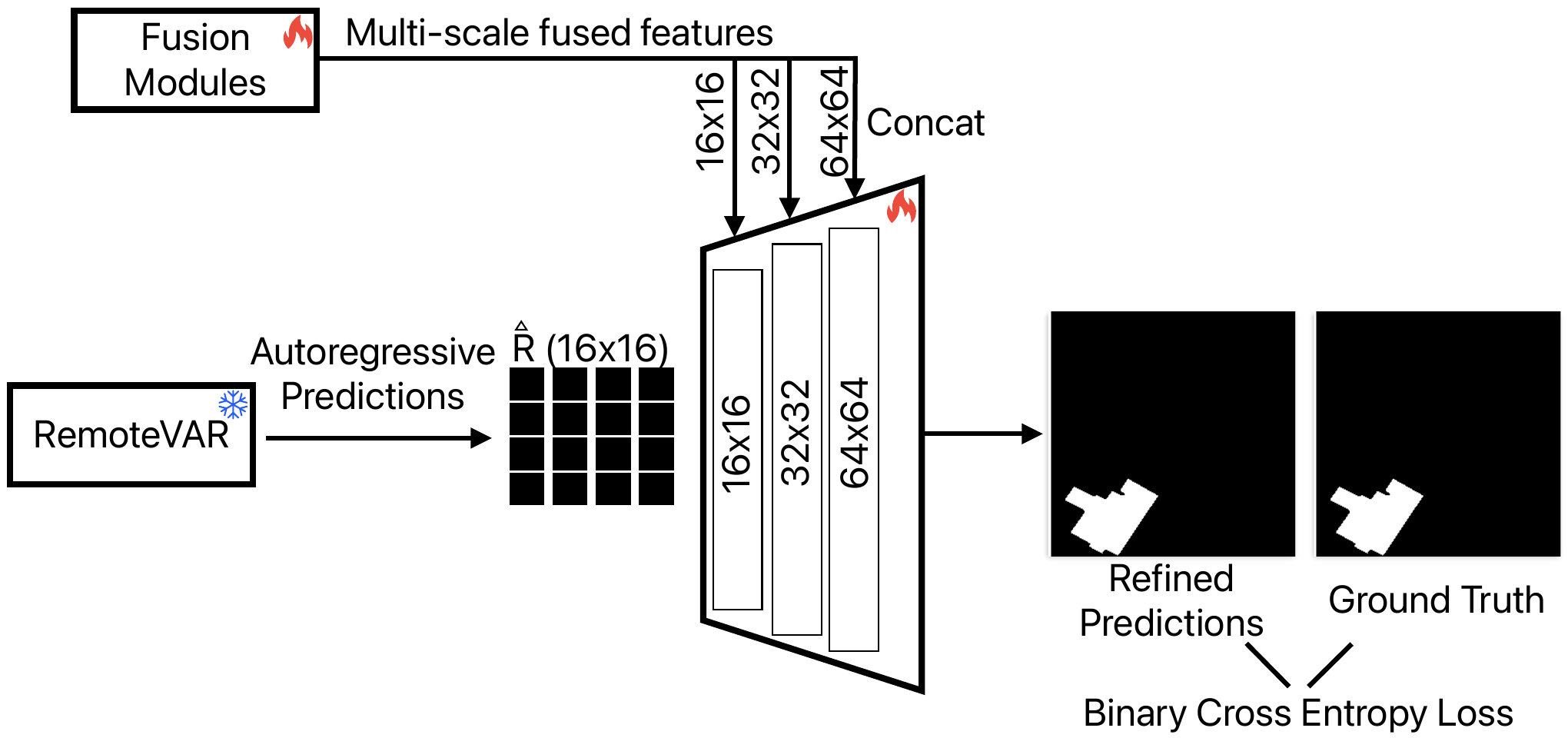}
    \caption{Decoder refinement procedure is illustrated.}
    \label{fig:inference}
\end{figure}

After autoregressive inference, the predicted mask tokens are mapped back through the codebook and decoded to pixel space by the VQ-VAE decoder. We then perform decoder refining to further sharpen boundaries and recover fine details: we augment the decoder with UNet-style skip connections, where the skip features come from our multi-scale fusion modules (rather than directly from the encoder), and fine-tune the decoder with a binary cross-entropy objective on the change map. This refinement improves boundary localization and small-object changes while keeping the autoregressive generator fixed. This process is illustrated in Fig. \ref{fig:inference}.

\begin{table*}
\begin{center}
\caption{Quantitative comparison on WHU-CD~\cite{whu} and LEVIR-CD~\cite{levir} using F1, IoU, and overall accuracy (OA).}
\resizebox{.6\linewidth}{!}{
\begin{tabular}
{@{\extracolsep{4pt}}c c c c c c c @{}}
\toprule
 & \multicolumn{3}{c}{WHU-CD \cite{whu}} & \multicolumn{3}{c}{LEVIR-CD \cite{levir}}\\
\cline{2-4} \cline{5-7} \\
Method & F1 (\(\uparrow\)) & IoU (\(\uparrow\)) & OA (\(\uparrow\)) & F1 (\(\uparrow\)) & IoU (\(\uparrow\)) & OA (\(\uparrow\)) \\
\midrule
SNUNet \cite{snunet} & 0.835 & 0.717 & 98.7 & 0.882 & 0.788 & 98.8\\
DT-SCN \cite{dtscn} & 0.914 & 0.842 & 99.3 & 0.877 & 0.781 & 98.8\\
STANet \cite{stanet} & 0.823 & 0.700 & 98.5 & 0.873 & 0.774 & 98.7\\
SeCo \cite{seco} & 0.883 & 0.790 & - & 0.881 & 0.787 & -\\
SaDL-CD \cite{sadlcd} & 0.909 & 0.833 & - & 0.899 & 0.818 & -\\
BIT \cite{bit} & 0.905 & 0.834 & 99.3 & 0.893 & 0.807 & 98.9\\
ChangeFormer \cite{changeformer} & 0.886 & 0.795 & 99.1 & 0.904 & 0.825 & 99.0\\
RSMamba \cite{rsmamba} & 0.927 & 0.865 & 99.4 & 0.897 & 0.814 & 98.9 \\
DDPM-CD \cite{ddpmcd} & 0.927 & 0.863 & 99.4 & 0.909 & 0.833 & 99.1\\
\midrule
RemoteVAR (Ours) & 0.930 & 0.870 & 99.4  & 0.910  &  0.834  & 99.1 \\
\bottomrule
\label{tab:quant_results}
\end{tabular}}
\end{center}
\end{table*}

\section{Experiments and Results}
\subsection{Implementation Details}
All images are resized to $256\times256$, and models are trained for 100 epochs with a batch size of 48 using AdamW (learning rate $1\times10^{-4}$). Weight decay is set to $1\times10^{-4}$ and annealed to 0 with a cosine schedule. The learning rate follows a cosine schedule with linear warmup, and gradients are clipped to a norm of 2.0. Mixed-precision (FP16) training is applied. To increase the number of training samples and promote token diversity, training is performed on the union of binary change detection datasets WHU-CD \cite{whu}, LEVIR-CD \cite{levir}, LEVIR-CD+ \cite{levir}, and S2Looking\cite{shen2021s2looking}. All experiments are run on 8 NVIDIA A5000 GPUs with 24\,GB memory each.

\subsection{Experimental setup.}
We evaluate RemoteVAR on two widely used building change detection benchmarks, LEVIR-CD~\cite{levir} and WHU-CD~\cite{whu}. LEVIR-CD contains VHR Google Earth image pairs with building changes, while WHU-CD provides large-scale aerial imagery collected at different years for building change detection. Following standard practice, we report performance on the official splits using F1 score, IoU, and overall pixel accuracy (OA) as the primary metrics, where F1/IoU emphasize the changed class under heavy class imbalance.

\begin{figure*}
    \centering
    \includegraphics[width=.9\linewidth]{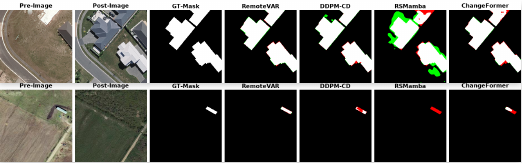}
    \caption{Qualitative prediction comparisons are shown for the WHU-CD \cite{whu} (top row) and LEVIR-CD \cite{levir} (bottom row) datasets. True positives are colored white, false positives green, and false negatives red.}
    \label{fig:comparison}
\end{figure*}

\subsection{Baselines}
We compare against representative diffusion-, Mamba/SSM-, transformer-, CNN-, and self-supervised approaches. DDPM-CD~\cite{ddpmcd} is a diffusion-feature-based change detector, RSMamba~\cite{rsmamba} uses a Siamese Mamba/SSM backbone, and ChangeFormer~\cite{changeformer} and BiT~\cite{bit} are transformer-based bi-temporal change detection models. For CNN baselines, SNUNet~\cite{snunet} and STANet~\cite{stanet} follow Siamese encoder--decoder designs with enhanced fusion/attention. We also include self-supervised pretraining baselines, SeCo~\cite{seco} and SaDL-CD~\cite{sadlcd}, which learn representations from unlabeled temporal imagery and are then fine-tuned for change detection.

\subsection{Results}
Quantitative results are reported in Table~\ref{tab:quant_results}. On WHU-CD, RemoteVAR achieves the best performance with $\mathrm{F1}=0.930$ and $\mathrm{IoU}=0.870$, slightly improving over strong baselines such as DDPM-CD~\cite{ddpmcd} ($0.927/0.863$) and RSMamba~\cite{rsmamba} ($0.927/0.865$). On LEVIR-CD, RemoteVAR remains competitive and reaches the top performance with $\mathrm{F1}=0.910$ and $\mathrm{IoU}=0.834$, marginally exceeding DDPM-CD ($0.909/0.833$) and outperforming the remaining transformer/CNN/self-supervised baselines by varying margins. Qualitative comparisons in Fig.~\ref{fig:comparison} further show that RemoteVAR produces cleaner masks and detects subtle structural changes with more accurate localization, which we attribute to its coarse-to-fine autoregressive prediction and progressive refinement across scales.

\subsection{Ablation Studies}
We analyze the contribution of each design choice on WHU-CD \cite{whu}, with results summarized in Table~\ref{ablation_compact}. ``No Cross-Att'' removes our cross-attention conditioning and yields a ControlVAR-style variant that conditions only through self-attention over discrete tokens; unlike the original ControlVAR~\cite{li2024controlvar} (single-condition), our setting is bi-conditioned on both pre- and post-image tokens. ``No Location Aware RGB Masks'' disables our RGB-based mask token conversion and uses binary mask tokens directly, which increases token imbalance and reduces mask token diversity. ``No TokRand'' removes the early-scale token randomization used during teacher forcing; this variant is directly related to exposure bias because the model is trained only on ground-truth coarse tokens and is not exposed to imperfect coarse predictions that arise at autoregressive inference. ``No DecRef'' disables the decoder refining stage and outputs the change map using the fixed VQ-VAE decoder (pure autoregressive prediction without refinement). Overall, each component provides complementary gains, and the full RemoteVAR achieves the best performance.

\begin{table}[ht]
\centering
\caption{Ablation results are presented in WHU-CD \cite{whu}.}
\setlength{\tabcolsep}{6pt}
\renewcommand{\arraystretch}{1.05}
\begin{tabular}{lccc}
\toprule
Variant & F1 & IoU & OA (\%) \\
\midrule
No Cross-Att (ControlVAR\cite{li2024controlvar}) & 0.145 & 0.078 & 96.0 \\
No Location Aware RGB Masks & 0.643 & 0.474 & 96.8 \\
No TokRand (exposure bias) & 0.877 & 0.781 & 99.0 \\
No DecRef (pure autoregressive) & 0.894 & 0.809 & 99.1 \\
\midrule
RemoteVAR & 0.930 & 0.870 & 99.4 \\
\bottomrule
\end{tabular}
\vspace{-5pt}
\label{ablation_compact}
\end{table}

\section{Conclusion}
We introduce RemoteVAR, a bi-temporal autoregressive change detection framework that generates multi-scale mask tokens in a coarse-to-fine manner. By combining discrete token generation with fused continuous pre/post features injected via cross-attention, RemoteVAR achieves accurate localization and strong performance on public benchmarks, matching or surpassing competitive Diffusion-, Transformer-, Mamba-, and CNN-based baselines. The results suggest that autoregressive modeling is a practical and effective alternative for dense remote sensing change detection.

\small
\bibliographystyle{IEEEtranN}
\bibliography{references}

@String(ICCV= {Int. Conf. Comput. Vis.})

@String(ICIP = {IEEE Int. Conf. Image Process.})

@String(ICCV  = {ICCV})

@String(ICIP  = {ICIP})

@INPROCEEDINGS{fcsiam,
  author={Caye Daudt, Rodrigo and Le Saux, Bertr and Boulch, Alexandre},
  booktitle={2018 25th IEEE International Conference on Image Processing (ICIP)}, 
  title={Fully Convolutional Siamese Networks for Change Detection}, 
  year={2018},
  volume={},
  number={},
  pages={4063-4067},
  doi={10.1109/ICIP.2018.8451652}}

@ARTICLE{snunet,
  author={Fang, Sheng and Li, Kaiyu and Shao, Jinyuan and Li, Zhe},
  journal={IEEE Geoscience and Remote Sensing Letters}, 
  title={SNUNet-CD: A Densely Connected Siamese Network for Change Detection of VHR Images}, 
  year={2022},
  volume={19},
  number={},
  pages={1-5},
  doi={10.1109/LGRS.2021.3056416}}

@INPROCEEDINGS{changeformer,
  author={Bandara, Wele Gedara Chaminda and Patel, Vishal M.},
  booktitle={IGARSS 2022 - 2022 IEEE International Geoscience and Remote Sensing Symposium}, 
  title={A Transformer-Based Siamese Network for Change Detection}, 
  year={2022},
  volume={},
  number={},
  pages={207-210},
  doi={10.1109/IGARSS46834.2022.9883686}}

@ARTICLE{dtscn,
  author={Liu, Yi and Pang, Chao and Zhan, Zongqian and Zhang, Xiaomeng and Yang, Xue},
  journal={IEEE Geoscience and Remote Sensing Letters}, 
  title={Building Change Detection for Remote Sensing Images Using a Dual-Task Constrained Deep Siamese Convolutional Network Model}, 
  year={2021},
  volume={18},
  number={5},
  pages={811-815},
  doi={10.1109/LGRS.2020.2988032}}

@article{stanet,
AUTHOR = {Chen, Hao and Shi, Zhenwei},
TITLE = {A Spatial-Temporal Attention-Based Method and a New Dataset for Remote Sensing Image Change Detection},
JOURNAL = {Remote Sensing},
VOLUME = {12},
YEAR = {2020},
NUMBER = {10},
ARTICLE-NUMBER = {1662},
URL = {https://www.mdpi.com/2072-4292/12/10/1662},
ISSN = {2072-4292},
DOI = {10.3390/rs12101662}
}

@article{ifnet,
title = {A deeply supervised image fusion network for change detection in high resolution bi-temporal remote sensing images},
journal = {ISPRS Journal of Photogrammetry and Remote Sensing},
volume = {166},
pages = {183-200},
year = {2020},
issn = {0924-2716},
doi = {https://doi.org/10.1016/j.isprsjprs.2020.06.003},
url = {https://www.sciencedirect.com/science/article/pii/S0924271620301532},
author = {Chenxiao Zhang and Peng Yue and Deodato Tapete and Liangcun Jiang and Boyi Shangguan and Li Huang and Guangchao Liu},
}

@ARTICLE{bit,
  author={Chen, Hao and Qi, Zipeng and Shi, Zhenwei},
  journal={IEEE Transactions on Geoscience and Remote Sensing}, 
  title={Remote Sensing Image Change Detection With Transformers}, 
  year={2022},
  volume={60},
  number={},
  pages={1-14},
  doi={10.1109/TGRS.2021.3095166}}

@misc{ddpmcd,
    title={DDPM-CD: Denoising Diffusion Probabilistic Models as Feature Extractors for Change Detection}, 
    author={Wele Gedara Chaminda Bandara and Nithin Gopalakrishnan Nair and Vishal M. Patel},
    year={2024},
    eprint={2206.11892},
    archivePrefix={arXiv},
    primaryClass={cs.CV},
    doi = {10.48550/ARXIV.2206.11892},
}

@Article{levir,
AUTHOR = {Chen, Hao and Shi, Zhenwei},
TITLE = {A Spatial-Temporal Attention-Based Method and a New Dataset for Remote Sensing Image Change Detection},
JOURNAL = {Remote Sensing},
VOLUME = {12},
YEAR = {2020},
NUMBER = {10},
ARTICLE-NUMBER = {1662},
URL = {https://www.mdpi.com/2072-4292/12/10/1662},
ISSN = {2072-4292},
DOI = {10.3390/rs12101662}
}

@ARTICLE{whu,
  author={Ji, Shunping and Wei, Shiqing and Lu, Meng},
  journal={IEEE Transactions on Geoscience and Remote Sensing}, 
  title={Fully Convolutional Networks for Multisource Building Extraction From an Open Aerial and Satellite Imagery Data Set}, 
  year={2019},
  volume={57},
  number={1},
  pages={574-586},
  doi={10.1109/TGRS.2018.2858817}}

@article{wen2024gcd,
  title={GCD-DDPM: A generative change detection model based on difference-feature-guided DDPM},
  author={Wen, Yihan and Ma, Xianping and Zhang, Xiaokang and Pun, Man-On},
  journal={IEEE Transactions on Geoscience and Remote Sensing},
  volume={62},
  pages={1--16},
  year={2024},
  publisher={IEEE}
}

@article{li2024controlvar,
  title={Controlvar: Exploring controllable visual autoregressive modeling},
  author={Li, Xiang and Qiu, Kai and Chen, Hao and Kuen, Jason and Lin, Zhe and Singh, Rita and Raj, Bhiksha},
  journal={arXiv preprint arXiv:2406.09750},
  year={2024}
}

@article{shen2021s2looking,
  title={S2Looking: A satellite side-looking dataset for building change detection},
  author={Shen, Li and Lu, Yao and Chen, Hao and Wei, Hao and Xie, Donghai and Yue, Jiabao and Chen, Rui and Lv, Shouye and Jiang, Bitao},
  journal={Remote Sensing},
  volume={13},
  year={2021},
  publisher={MDPI}
}

@article{zhang2023cmx,
  title={CMX: Cross-modal fusion for RGB-X semantic segmentation with transformers},
  author={Zhang, Jiaming and Liu, Huayao and Yang, Kailun and Hu, Xinxin and Liu, Ruiping and Stiefelhagen, Rainer},
  journal={IEEE Transactions on intelligent transportation systems},
  volume={24},
  number={12},
  pages={14679--14694},
  year={2023},
  publisher={IEEE}
}

@ARTICLE{cd1,
  author={Khelifi, Lazhar and Mignotte, Max},
  journal={IEEE Access}, 
  title={Deep Learning for Change Detection in Remote Sensing Images: Comprehensive Review and Meta-Analysis}, 
  year={2020},
  volume={8},
  number={},
  pages={126385-126400},
  doi={10.1109/ACCESS.2020.3008036}}

@article{climateshift_cd1,
author = {Liu, Qunqun and Wan, Shiquan and Gu, Bin},
year = {2016},
month = {01},
pages = {1-10},
title = {A Review of the Detection Methods for Climate Regime Shifts},
volume = {2016},
journal = {Discrete Dynamics in Nature and Society},
doi = {10.1155/2016/3536183}
}

@ARTICLE{disaster_cd1,
  author={Hamidi, Ebrahim and Peter, Brad G. and Muñoz, David F. and Moftakhari, Hamed and Moradkhani, Hamid},
  journal={IEEE Transactions on Geoscience and Remote Sensing}, 
  title={Fast Flood Extent Monitoring With SAR Change Detection Using Google Earth Engine}, 
  year={2023},
  volume={61},
  number={},
  pages={1-19},
  doi={10.1109/TGRS.2023.3240097}}

@article{disaster_cd2,
title = {Cross-modal change detection flood extraction based on convolutional neural network},
journal = {International Journal of Applied Earth Observation and Geoinformation},
volume = {117},
pages = {103197},
year = {2023},
issn = {1569-8432},
doi = {https://doi.org/10.1016/j.jag.2023.103197},
url = {https://www.sciencedirect.com/science/article/pii/S1569843223000195},
author = {Xiaoning He and Shuangcheng Zhang and Bowei Xue and Tong Zhao and Tong Wu}
}

@article{policy_cd1,
title = {A Comparison of Four Algorithms for Change Detection in an Urban Environment},
journal = {Remote Sensing of Environment},
volume = {63},
number = {2},
pages = {95-100},
year = {1998},
issn = {0034-4257},
doi = {https://doi.org/10.1016/S0034-4257(97)00112-0},
url = {https://www.sciencedirect.com/science/article/pii/S0034425797001120},
author = {Merrill K Ridd and Jiajun Liu},
}

@article{land_cd1,
author = {Kaur, Ravneet and Tiwari, Reet and Maini, Raman and Singh, Sartajvir},
year = {2023},
month = {04},
pages = {28},
title = {A Framework for Crop Yield Estimation and Change Detection Using Image Fusion of Microwave and Optical Satellite Dataset},
volume = {6},
journal = {Quaternary},
doi = {10.3390/quat6020028}
}

@article{land_cd2,
author = {Coops, Nicholas and Tompalski, Piotr and Goodbody, Tristan and Achim, Alexis and Mulverhill, Christopher},
year = {2022},
month = {05},
pages = {1-19},
title = {Framework for near real-time forest inventory using multi source remote sensing data},
journal = {Forestry},
doi = {10.1093/forestry/cpac015}
}

@article{military_cd1,
author = {Tueller, Paul and Ramsey, Robert and Frank, Thomas and Washington-Allen, Robert and Tweddale, Scott},
year = {1998},
month = {12},
pages = {74},
title = {Emerging and Contemporary Technologies in Remote Sensing for Ecosystem Assessment and Change Detection on Military Reservations}
}

@ARTICLE{resolution_challenge1,
  author={Benediktsson, Jon Atli and Chanussot, Jocelyn and Moon, Wooil M.},
  journal={Proceedings of the IEEE}, 
  title={Very High-Resolution Remote Sensing: Challenges and Opportunities [Point of View]}, 
  year={2012},
  volume={100},
  number={6},
  pages={1907-1910},
  doi={10.1109/JPROC.2012.2190811}}

@misc{noise_challenge1,
      title={A Comparative Study of Removal Noise from Remote Sensing Image}, 
      author={Salem Saleh Al-amri and N. V. Kalyankar and S. D. Khamitkar},
      year={2010},
      eprint={1002.1148},
      archivePrefix={arXiv},
      primaryClass={cs.CV}
}

@ARTICLE{noise_challenge2,
  author={Landgrebe, David A. and Malaret, Erick},
  journal={IEEE Transactions on Geoscience and Remote Sensing}, 
  title={Noise in Remote-Sensing Systems: The Effect on Classification Error}, 
  year={1986},
  volume={GE-24},
  number={2},
  pages={294-300},
  doi={10.1109/TGRS.1986.289648}}

@ARTICLE{registration_challenge1,
  author={Inglada, Jordi and Muron, Vincent and Pichard, Damien and Feuvrier, Thomas},
  journal={IEEE Transactions on Geoscience and Remote Sensing}, 
  title={Analysis of Artifacts in Subpixel Remote Sensing Image Registration}, 
  year={2007},
  volume={45},
  number={1},
  pages={254-264},
  doi={10.1109/TGRS.2006.882262}}

@ARTICLE{registration_challenge2,
  author={Bentoutou, Y. and Taleb, N. and Kpalma, K. and Ronsin, J.},
  journal={IEEE Transactions on Geoscience and Remote Sensing}, 
  title={An automatic image registration for applications in remote sensing}, 
  year={2005},
  volume={43},
  number={9},
  pages={2127-2137},
  doi={10.1109/TGRS.2005.853187}}

@ARTICLE{illumination_challenge1,
  author={Wan, Xue and Liu, Jianguo and Li, Shengyang and Dawson, John and Yan, Hongshi},
  journal={IEEE Transactions on Geoscience and Remote Sensing}, 
  title={An Illumination-Invariant Change Detection Method Based on Disparity Saliency Map for Multitemporal Optical Remotely Sensed Images}, 
  year={2019},
  volume={57},
  number={3},
  pages={1311-1324},
  doi={10.1109/TGRS.2018.2865961}}

@article{illumination_challenge2,
  title={Illumination and Contrast Balancing for Remote Sensing Images},
  author={Jun Liu and Xing Wang and Min Chen and Shuguang Liu and Zhenfeng Shao and Xiran Zhou and Ping Liu},
  journal={Remote. Sens.},
  year={2014},
  volume={6},
  pages={1102-1123},
  url={https://api.semanticscholar.org/CorpusID:16993816}
}

@InProceedings{seco,
    author    = {Ma\~nas, Oscar and Lacoste, Alexandre and Gir\'o-i-Nieto, Xavier and Vazquez, David and Rodr{\'\i}guez, Pau},
    title     = {Seasonal Contrast: Unsupervised Pre-Training From Uncurated Remote Sensing Data},
    booktitle = {Proceedings of the IEEE/CVF International Conference on Computer Vision (ICCV)},
    month     = {October},
    year      = {2021},
    pages     = {9414-9423}
}

@ARTICLE{sadlcd,
  author={Chen, Hao and Li, Wenyuan and Chen, Song and Shi, Zhenwei},
  journal={IEEE Transactions on Geoscience and Remote Sensing}, 
  title={Semantic-Aware Dense Representation Learning for Remote Sensing Image Change Detection}, 
  year={2022},
  volume={60},
  number={},
  pages={1-18},
  doi={10.1109/TGRS.2022.3203769}}

@misc{rsmamba,
      title={RS-Mamba for Large Remote Sensing Image Dense Prediction}, 
      author={Sijie Zhao and Hao Chen and Xueliang Zhang and Pengfeng Xiao and Lei Bai and Wanli Ouyang},
      year={2024},
      eprint={2404.02668},
      archivePrefix={arXiv},
      primaryClass={id='cs.CV' full_name='Computer Vision and Pattern Recognition' is_active=True alt_name=None in_archive='cs' is_general=False description='Covers image processing, computer vision, pattern recognition, and scene understanding. Roughly includes material in ACM Subject Classes I.2.10, I.4, and I.5.'}
}

@article{tian2024visual,
  title={Visual autoregressive modeling: Scalable image generation via next-scale prediction},
  author={Tian, Keyu and Jiang, Yi and Yuan, Zehuan and Peng, Bingyue and Wang, Liwei},
  journal={Advances in neural information processing systems},
  volume={37},
  pages={84839--84865},
  year={2024}
}

\end{document}